\documentclass[runningheads]{llncs}

\usepackage[T1]{fontenc}
\usepackage{amsmath,amssymb,amsfonts}
\usepackage{graphicx,verbatim}
\usepackage{algorithm}
\usepackage{hyperref}
\usepackage{cleveref}
\usepackage{cite}
\usepackage{algpseudocode}
\usepackage{nicefrac}
\usepackage{comment}
\usepackage{microtype} 
\usepackage{textcomp}
\usepackage{cleveref}
\usepackage{multirow}
\usepackage{tikz}
\usetikzlibrary{intersections}
\usepackage{float}
\usepackage{url}
\usepackage{pifont}
\usepackage{tabularx}

\usepackage{enumitem}
\usepackage{booktabs}
\usepackage{stfloats} 
\usepackage{subcaption}
\usepackage{comment}
\usepackage{stfloats} 
\usepackage[dvipsnames]{xcolor}
\newcommand{\edit}[1]{\textcolor{ForestGreen}{#1}}
\renewcommand{\edit}[1]{#1}
\usetikzlibrary{positioning, shapes.multipart, fit, calc, arrows.meta, decorations.pathreplacing}
\usepackage{pgfplots}
\usepgfplotslibrary{fillbetween}
\pgfplotsset{compat=1.18}
\usepgfplotslibrary{colorbrewer}
\pgfplotsset{cycle list/Set1-6} 
\definecolor{majorColor}{RGB}{31,119,180} 
\definecolor{minorColor}{RGB}{214,39,40}  

\begin{document}
\title{S-CEReBrO: Breaking the Memory Barrier in Continuous EEG Monitoring\thanks{This is the pre-rebuttal version of a paper accepted at MICCAI 2026.}}
%
\author{Glenn Anta Bucagu\inst{1} \and
Thorir Mar Ingolfsson\inst{1} \and
Yawei Li\inst{1,2} \and
Luca Benini\inst{1,3}}

\authorrunning{G. Bucagu et al.}
%
\institute{ETH Zurich, Zurich, Switzerland \and
Nanyang Technological University, Singapore \and
University of Bologna, Bologna, Italy \\
\email{gbucagu@iis.ee.ethz.ch}}

\maketitle              

\begin{abstract}
Foundation models offer a promising paradigm for Electroencephalography (EEG) analysis, leveraging generalizable representations from vast unlabeled datasets. Yet, Transformer-based architectures face a critical bottleneck: global attention mechanisms couple the attention memory state to the signal duration, causing memory overflow during continuous monitoring. To address this, we introduce \textbf{S-CEReBrO} (Streaming CEReBrO), an evolution of the CEReBrO architecture designed for continuous monitoring. Our novel \textit{Windowed Alternating Attention} mechanism factorizes attention computation into fixed-size spatiotemporal windows, so that
under streaming, only the active window remains resident and the
attention state is bounded independently of signal duration. \edit{Empirical scaling analysis shows that windowed alternating attention can process signals \textbf{100$\times$} longer than full self-attention and \textbf{3$\times$} longer than low-rank linear attention. Compared to low-rank linear attention on long contexts, windowed alternating attention requires \textbf{55\%} of the memory while increasing inference throughput by \textbf{2.1$\times$}}. Pre-trained on $>$25,000 hours of recordings from $>$12,000 subjects, S-CEReBrO achieves state-of-the-art performance on \textbf{7 of 11} downstream tasks, with up to \textbf{60}\% fewer parameters. This work represents a significant step toward the realization of efficient, generalizable, and continuous EEG monitoring. \edit{An accompanying code repository is available \footnote{\url{https://github.com/pulp-bio/biofoundation}}}.


\end{abstract}    
\section{Introduction}
\label{sec:intro}

Electroencephalography (EEG) non-invasively records the brain's electrical activity by capturing voltage fluctuations at the scalp. As the field shifts toward continuous, long-term monitoring for applications like seizure detection, the resulting data, structured as $C$ channels $\times$ $T$ timesteps, poses a scaling challenge. While hardware limits $C$ to a fixed range (typically 2--64), $T$ grows indefinitely during prolonged observation. To process this high-dimensional data, EEG foundation models have emerged as a powerful alternative to supervised methods, leveraging large-scale pre-training to learn general representations. Although the Transformer architecture currently dominates this field~\cite{LaBraM, BIOT, CBraMod} and shows promise for resource-constrained inference~\cite{CEReBrO}, a critical memory bottleneck remains. Wearable devices cannot usually store the whole history of Transformer key and value (KV) states for large $T$. This memory-bound limitation prevents high-performance foundation models from sustaining the processing required for continuous EEG monitoring.

To bridge this gap, we introduce \textbf{S-CEReBrO} (Streaming CEReBrO), an evolution of the CEReBrO architecture~\cite{CEReBrO}. S-CEReBrO utilizes a novel \textit{Windowed Alternating Attention} mechanism that interleaves spatial and temporal attention blocks, restricting interactions to local neighborhoods in time and space. This allows our model to process long signals while maintaining a constant resident memory. We summarize our contributions as follows:
\begin{itemize}[leftmargin=*]
    \item \textbf{Locality-Driven Attention:} We introduce a Windowed Alternating Attention that maintains a constant peak resident memory by evicting stale tokens from the computational buffer. While total computational complexity scales linearly ($\mathcal{O}(CT)$) to process the full signal, our approach eliminates the cumulative memory overflow inherent to global attention. \edit{This mechanism can handle 14-hour continuous monitoring, with 100$\times$ the capacity of full self-attention, while needing 55\% of the memory of low-rank linear attention and achieving 2.1$\times$ higher inference throughput.}

    \item \textbf{Cross-Task SoTA Performance:} We demonstrate that global attention is often redundant for EEG analysis. S-CEReBrO outperforms global attention baselines on 7 of 11 public benchmarks, spanning clinical diagnostics, BCI, and state monitoring. It achieves these results while using up to 60\% fewer parameters than existing models.

\end{itemize}
    
\begin{table}[!t]
  \centering
  \caption{Comparison of Transformer-based EEG foundation models. S-CEReBrO is the only architecture that combines strictly linear attention complexity with a constant resident memory footprint relative to signal duration ($T$).}
  \label{tab:model_summary}
  \small
  \setlength{\tabcolsep}{5pt}
  \renewcommand{\arraystretch}{1.2}
  
  \resizebox{\columnwidth}{!}{%
  \begin{tabular}{l c l l c}
    \toprule
    \textbf{Model} & \textbf{Params (M)} & \textbf{Attention Mechanism} & \textbf{Attention Complexity} & \textbf{Peak Memory} \\
    \midrule
    BIOT~\cite{BIOT}       & 3.2      & Linear                 & $\mathcal{O}(CT)$         & $\mathcal{O}(T)$   \\
    CBraMod~\cite{CBraMod} & 4.0      & Criss-Cross            & $\mathcal{O}(CT(C + T))$  & $\mathcal{O}(T)$   \\
    LaBraM~\cite{LaBraM}   & 5.8--369 & Full Self-Attention  & $\mathcal{O}((CT)^2)$     & $\mathcal{O}(T)$ \\
    LUNA~\cite{LUNA}       & 7.0--311 & Perceiver              & $\mathcal{O}(T(C + T))$   & $\mathcal{O}(T)$   \\
    CEReBrO~\cite{CEReBrO} & 2.4 & Alternating       & $\mathcal{O}(CT(C + T))$  & $\mathcal{O}(T)$   \\
    \textbf{S-CEReBrO (Ours)} & \textbf{2.4} & \textbf{Windowed Alternating} & \textbf{$\mathcal{O}(CT)$} & \textbf{$\mathcal{O}(1)$} \\
    \bottomrule
  \end{tabular}%
  }

  \begin{minipage}{\columnwidth}
    \scriptsize
    \textbf{Peak Memory}: Peak resident state growth in $T$ during streaming. Bounded token dependencies allow eviction, achieving $\mathcal{O}(1)$ memory. Otherwise, the full sequence must be retained, resulting in $\mathcal{O}(T)$. \\
  \end{minipage}
\end{table}

\section{Related Works}
\label{sec:related_works}

Although Transformer-based EEG foundation models excel at learning generalized representations, the high computational cost of their global attention mechanisms prevents them from processing the long contexts required for continuous monitoring. Comparisons of the current state-of-the-art are available in \Cref{tab:model_summary}.

Models like LaBraM~\cite{LaBraM} achieve high performance but inherit the quadratic complexity $\mathcal{O}((CT)^2)$ of vanilla Transformers. BIOT~\cite{BIOT} approximates attention via low-rank projections~\cite{wang2020linformer} to achieve $\mathcal{O}(CT)$ complexity. CBraMod~\cite{CBraMod} employs Criss-Cross attention to separate temporal and channel attention across different heads, with $\mathcal{O}(CT(C + T))$ complexity. LUNA~\cite{LUNA} maps high-dimensional EEG inputs onto a fixed-size set of latent variables, yielding $\mathcal{O}(T(C + T))$ complexity. CEReBrO~\cite{CEReBrO} uses alternating attention to separate spatial and temporal attention in separate layers. However, all these models maintain a global receptive field, meaning their resident memory scales with $T$. In contrast, S-CEReBrO enforces algorithmic locality through \textit{Windowed Alternating Attention}, ensuring a resident memory state that is independent from $T$. While windowed attention mechanisms have been popularized by Large Language Models to handle long context~\cite{longformer}, these methods are not directly transferable to neural signals. Unlike text, which adheres to a 1D linear grammar, EEG signals exhibit a complex 2D non-linear structure. 

\section{Method}
\label{sec:method}
\begin{figure*}[!t] 
    \centering
    \includegraphics[width=0.8\textwidth]{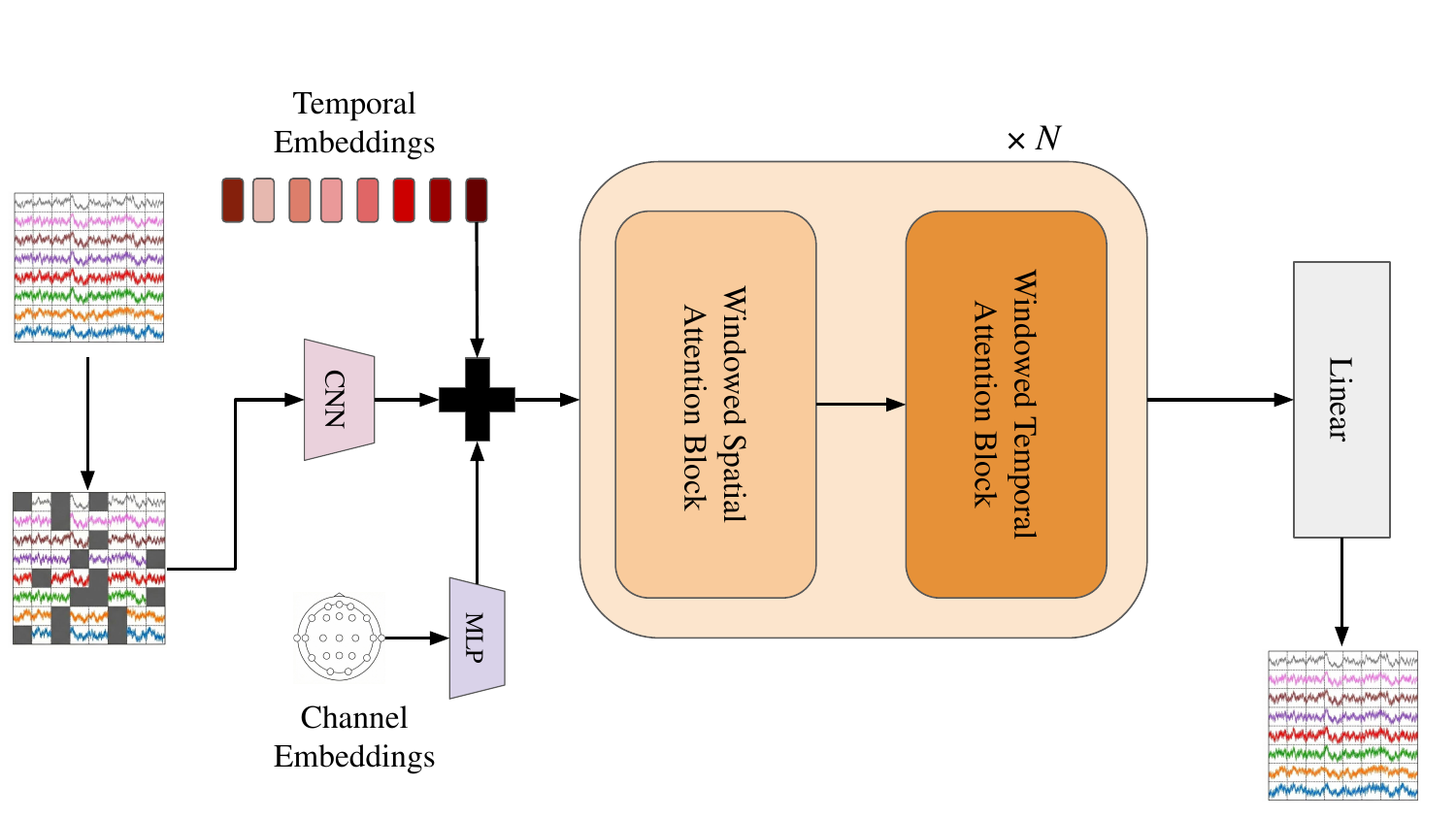}
    \caption{Overview of the S-CEReBrO architecture and masked autoencoding pipeline, featuring $N$ composite attention blocks, where each comprises a WSA and a WTA layer.}
    \label{fig:model_architecture}
\end{figure*}

S-CEReBrO reconciles the long context requirements of continuous EEG monitoring with the constraints of real-world hardware. We use a novel \textit{Windowed Alternating Attention} mechanism with linear complexity $O(CT)$ and a constant resident memory. S-CEReBrO learns useful representations via masked autoencoding. ~\Cref{fig:model_architecture} summarizes the pipeline.  We detail each component below.

\subsection{Spatiotemporal Tokenization}
Given a raw EEG recording $\mathbf{X} \in \mathbb{R}^{C \times T}$, we tokenize the signal into a sequence of patch embeddings:

\begin{enumerate}[leftmargin=*]
    \item \textbf{Patching:} We apply a sliding window of length $L$ and stride $S$ to each channel, yielding a patch tensor $\mathbf{P} \in \mathbb{R}^{C \times N_p \times L}$, where $N_p = \lfloor \frac{T-L}{S} \rfloor + 1$~\cite{patchtst}.
    \item \textbf{Feature Projection:} A shared convolutional encoder projects each raw patch $\mathbf{P}_{c,i}$ into a latent embedding $\mathbf{E}_{c,i} \in \mathbb{R}^{d_e}$.
    \item \textbf{Spatiotemporal Embeddings:} We preserve spatiotemporal topology using two learnable embeddings: an index-based matrix $\mathbf{W}_{\text{pos}} \in \mathbb{R}^{N_p \times d_e}$ for temporal order, and a continuous spatial embedding $\mathbf{W}_{\text{chan},c} \in \mathbb{R}^{d_e}$ obtained by mapping 3D electrode coordinates $(x, y, z)$  via a shared MLP~\cite{LUNA}.
\end{enumerate}

The final input tensor to the Transformer, $\mathbf{H}^{(0)} \in \mathbb{R}^{C \times N_p \times d_e}$, is computed as:
\begin{equation}
    \mathbf{H}^{(0)}_{c,i} = \mathbf{E}_{c,i} + \mathbf{W}_{\text{pos},i} + \mathbf{W}_{\text{chan},c}
\end{equation}
where $\mathbf{E}_{c,i}$ is the projected patch embedding, $\mathbf{W}_{\text{pos},i}$ is the temporal embedding for index $i$, and $\mathbf{W}_{\text{chan},c}$ is the spatial embedding for channel $c$. 

\subsection{Windowed Temporal Attention (WTA)}
WTA layers model temporal dependencies independently for each channel $c$. To maintain a constant resident memory, each query $\mathbf{q}_{c,i}$ attends only to keys within a local neighborhood $\mathcal{N}_t$ of size $w_t$. 
\begin{equation}
    \text{WTA}(\mathbf{H}_{c,i}) = \text{Softmax}\left( \frac{\mathbf{q}_{c,i} (\mathbf{K}_{c, \mathcal{N}_t(i)})^T}{\sqrt{\frac{d_e}{h}}} \right) \mathbf{V}_{c, \mathcal{N}_t(i)}
\end{equation}
where the neighborhood $\mathcal{N}_t(i, l)$ for a time step $i$ at layer $l$ is defined as the set of valid indices:

\begin{equation}
    \mathcal{N}_t(i, l) = \{ k \in [0, N_p-1] \mid k = i + (j \cdot d_l + s_l)\}\footnote{We follow 0-indexed notation used in standard deep learning frameworks.} 
\end{equation}
for $j \in [-\lfloor \frac{w_t-1}{2} \rfloor, \dots, \lceil \frac{w_t-1}{2} \rceil]$. Layer-specific temporal dilation ($d_l$) and shift factors ($s_l$) expand the receptive field at constant attention cost and fixed, 
$T$-independent buffer size. Although individual WTA windows remain local, stacking them yields a larger effective timespan, covering the full pre-training window. WSA layers, acting at a fixed time index, ensure each WTA layer integrates channel-mixed rather than single-channel features.

\subsection{Windowed Spatial Attention (WSA)}
WSA layers facilitate cross-channel information exchange at a fixed time index $i$. Attention computation is restricted to a local $w_s$-sized neighborhood $\mathcal{N}_s$. 
\begin{equation}
    \text{WSA}(\mathbf{H}_{c,i}) = \text{Softmax}\left( \frac{\mathbf{q}_{c,i} (\mathbf{K}_{\mathcal{N}_s(c, l), i})^T}{\sqrt{\frac{d_e}{h}}} \right) \mathbf{V}_{\mathcal{N}_s(c, l), i}
\end{equation}
where the neighborhood $\mathcal{N}_s(c, l)$ for a channel $c$ at layer $l$ is defined as the set of valid indices:

\begin{equation}
    \mathcal{N}_s(c, l) = \{ n \in [0, C-1] \mid n = c + (m \cdot d_l + s_l) \}
\end{equation}
for $m \in [-\lfloor \frac{w_s-1}{2} \rfloor, \dots, \lceil \frac{w_s-1}{2} \rceil]$. Layer-specific dilation ($d_l$) and shift factors ($s_l$) stagger the receptive fields, enabling distant but physiologically related channels to interact across the model's depth. This design efficiently approximates global spatial relationships without the high memory overhead of distance-based maps or dynamic clusters. Beyond substantial efficiency gains (see \Cref{subsec:attention_efficiency}), our attention mechanism introduces a beneficial spatiotemporal prior that translates into superior performance on several downstream EEG tasks (\Cref{subsec:downstream_tasks}). Key hyperparameters are presented in \Cref{tab:hyperparameters}. 


\begin{table}[!t]
\centering
\caption{Key Hyperparameters for S-CEReBrO.}
\label{tab:hyperparameters}
\begin{tabular}{@{}ll@{}}
\toprule
\textbf{Hyperparameter} & \textbf{Value} \\ \midrule
Spatial Window Size ($w_s$) & min($C$, 7) \\
Temporal Window Size ($w_t$) & min($N_p$, 5) \\
Dilation Factors ($d_l$) & $\{1, 2, 4\}$ \\
Shift Factors ($s_l$) & $\{0, 1, -1, 2, -2\}$ \\
Number of Attention Block Pairs ($N$) & 3 \\
Embedding Dimension ($d_{e}$) & 180 \\
Number of Attention Heads ($h$) & 5 \\ \bottomrule
\end{tabular}
\end{table}

\subsection{Algorithmic Distinction and Computational Complexity}

The difference between windowed alternating attention and standard alternating attention lies in the \textit{management of the computational state}. Alternating attention~\cite{CEReBrO} requires retaining the key-value states for all $N_p$ temporal positions and $C$ spatial positions, leading to a $\mathcal{O}(N_p) \propto \mathcal{O}(T)$ resident memory footprint that eventually exceeds hardware limits as $T \to \infty$. Its total computational complexity is $\mathcal{O}(C N_p (C + N_p)) \propto \mathcal{O}(CT(C + T))$. In contrast, Windowed Alternating Attention enforces algorithmic locality. By evicting stale tokens outside the fixed neighborhoods $\mathcal{N}_t(i, l), \mathcal{N}_s(c, l)$ from the computational buffer, the resident memory requirement remains constant relative to $T$. Information from evicted tokens persists in the retained representations, which have already aggregated their neighborhoods through the WTA layers. Since $w_t$ and $w_s$ are hyperparameters fixed at design time, the total computational complexity is 
$\mathcal{O}(C N_p (w_t + w_s)) \propto \mathcal{O}(CT)$. Current state-of-the-art EEG foundation models are typically forced into arbitrary fixed-window chunking to mitigate memory bottlenecks, which treats the EEG signal as a series of isolated, independent segments. In contrast, S-CEReBrO employs First In First Out (FIFO)-based streaming to maintain a continuous, stateful receptive field. This approach preserves temporal dependencies across chunk boundaries, which fixed-window segmentation truncates.

\subsection{Pre-training}
Following~\cite{CEReBrO}, we employ Masked Autoencoding (MAE) for pre-training. Input sequences are tokenized into patches $\mathbf{P}$, with a fixed percentage randomly replaced by a shared learnable [MASK] token. S-CEReBrO processes these tokens, and a linear projection layer reconstructs the original patches as $\hat{\mathbf{P}}$. The pre-training objective minimizes the weighted sum of reconstruction losses for masked ($\mathcal{M}$) and visible ($\overline{\mathcal{M}}$) positions:

\begin{equation}
    \mathcal{L} = \mathcal{L}_{\text{m}} + \alpha\mathcal{L}_{\text{v}}, \text{ where } \mathcal{L}_{\{\text{m,v}\}} = \frac{1}{|\mathcal{K}|} \sum_{(c,i) \in \mathcal{K}} \| \mathbf{P}_{c,i} - \hat{\mathbf{P}}_{c,i} \|_2^2 \text{ for } \mathcal{K} \in \{\mathcal{M}, \overline{\mathcal{M}}\}
\end{equation}
\section{Experiments} \label{sec:experiments}

\subsection{Pre-training Corpus}
\label{subsec:pretraining_corpus}
S-CEReBrO is pre-trained on over 25,000 hours of EEG from more than 12,000 subjects, aggregating diverse clinical (TUEG~\cite{obeid2016temple}) and consumer-grade datasets (SEED series~\cite{SEED,SEED-IV,SEED-FRA}, BOAS~\cite{BOAS}, SleepEDFx~\cite{SleepEDFx}, BCI-NER~\cite{BCI-NER}, GWD~\cite{Kobler2018}). This corpus spans heterogeneous montages (2--62 channels) and referencing schemes, ensuring robustness across recording conditions. All signals are resampled to 200 Hz and processed in 30-second windows. To prevent downstream data leakage, subjects from the TUAB dataset are strictly excluded.

\subsection{Downstream Tasks}
\label{subsec:downstream_tasks}
To evaluate S-CEReBrO, we selected 11 downstream benchmarks spanning clinical neurology (seizure detection, sleep staging) and consumer BCI (motor imagery, emotion recognition). Summarized in Table \ref{tab:downstream_tasks}, these tasks comprise binary, multi-class, and regression objectives with heterogeneous sampling rates (128–1000 Hz), montages (6–64 channels), and durations (4–30 s). We employ subject-stratified splits to ensure zero overlap between sets. All benchmarks utilize subjects and channel configurations unseen during pre-training.

\subsection{Baselines and Performance Metrics}
\label{subsec:baselines_metrics}

We compare S-CEReBrO against leading non-foundational architectures (EEGNet~\cite{EEGNet}, EEGConformer~\cite{EEGConformer}, SPaRCNet~\cite{SPaRCNet}, CNN-Transformer~\cite{CNN-Transformer}, ST-Transformer \cite{ST-Transformer}, ContraWR~\cite{ContraWR} and FFCL~\cite{FFCL}). To reflect practical deployability on resource-constrained hardware, we restrict our EEG foundation model baselines to the highest performing models with fewer than 10M parameters. These are BIOT~\cite{BIOT}, LaBraM-Base~\cite{LaBraM} and CBraMod~\cite{CBraMod}. Model checkpoints are selected based on the optimal validation score. We use AUROC for binary classification, weighted F1-score for multi-class classification, and NRMSE for regression.

\begin{table}[!t]
\caption{Overview of downstream EEG decoding tasks and datasets used for fine-tuning.}
    \label{tab:downstream_tasks}
    \centering
    \resizebox{\textwidth}{!}{%
    \begin{tabular}{llcccccl}
        \toprule
        \textbf{Decoding Task} & \textbf{Dataset} & \textbf{Frequency(Hz)} & \textbf{Channels} & \textbf{Duration(s)} & \textbf{Subjects} & \textbf{ Samples} & \textbf{Objective} \\
        \midrule
        Seizure Detection & CHB-MIT~\cite{CHB-MIT} & 256 & 16 & 10 & 22 & 326,993 & 2-class \\
         & Neonate~\cite{Neonate} & 256 & 19 & 5 & 79 & 80,458 & 2-class \\
        Motor Imagery Classification & PhysioNet-MI~\cite{PhysioNet-MI} & 160 & 64 & 4 & 109 & 9,837 & 4-class \\
         & SHU-MI~\cite{SHU-MI} & 250 & 32 & 4 & 25 & 11,988 & 2-class \\
        Abnormal Classification & TUAB~\cite{obeid2016temple} & 250 & 22 & 10 & 2,383 & 409,455 & 2-class \\
        Mental Workload Classification & STEW~\cite{STEW} & 128 & 14 & 4 & 48 & 6,660 & 3-class \\
        Vigilance Estimation & SEED-VIG~\cite{SEED-VIG} & 200 & 17 & 8 & 23 & 20,355 & Regression \\
        Sleep Stage Classification & ISRUC~\cite{ISRUC} & 200 & 6 & 30 & 100 & 89,240 & 5-class \\
        Mental Disorder Diagnosis & Mumtaz~\cite{Mumtaz2016} & 256 & 19 & 5 & 62 & 7,143 & 2-class \\
        Mental Stress Detection & MentalArithmetic~\cite{MentalArithmetic} & 500 & 20 & 5 & 36 & 1,707 & 2-class \\
        Emotion Recognition & SEED-V~\cite{SEED-V} & 1000 & 62 & 4 & 20 & 117,744 & 5-class \\
        \bottomrule
    \end{tabular}%
    }
\end{table}

\subsection{Cross-Task Evaluation}
\label{subsec:results}

As detailed in \Cref{tab:main_results}, S-CEReBrO achieves state-of-the-art performance on 7 of 11 tasks. It outperforms both specialized baselines and larger foundation models despite its compact 2.4M parameter count. The most significant gains occur in clinical and behavioral monitoring. Specifically, in seizure detection (CHB-MIT, Neonate) and mental workload assessment (STEW, SEED-VIG), S-CEReBrO excels. These are popular use cases for continuous monitoring, where model efficiency and accuracy are paramount. \Cref{tab:stew_seedvig} confirms that the localized inductive bias of S-CEReBrO is particularly effective for cognitive monitoring. On the STEW and SEED-VIG benchmarks, S-CEReBrO achieves state-of-the-art performance, surpassing every model including its global counterpart (CEReBrO).

\begin{table*}[!hbtp]
    \centering
     \caption{Comparison of S-CEReBrO against state-of-the-art baselines across downstream tasks. We report mean $\pm$ standard deviation across five random seeds.}
     \label{tab:main_results}
    \resizebox{\textwidth}{!}{%
    \begin{tabular}{l l c | c c c | c}
        \toprule
        \textbf{Dataset} & \textbf{Metric} & \textbf{Best Supervised} & \textbf{BIOT}~\cite{BIOT} & \textbf{CBraMod}~\cite{CBraMod} & \textbf{LaBraM}~\cite{LaBraM} & \textbf{S-CEReBrO (Ours)} \\
        & & \textbf{Model} & (3.2M) & (4.0M) & (5.8M) & (2.4M) \\
        \midrule
        CHB-MIT    & AUROC $\uparrow$ & 86.62 $\pm$ 0.82 & 80.84 $\pm$ 4.43 & 84.55 $\pm$ 0.74 & 79.70 $\pm$ 5.64 & \textbf{87.45 $\pm$ 1.93} \\
        Neonate    & AUROC $\uparrow$ & 81.45 $\pm$ 4.96 & 79.78 $\pm$ 5.47 & 76.63 $\pm$ 6.58 & 79.78 $\pm$ 3.30 & \textbf{85.76 $\pm$ 2.49}  \\
        PhysioNet-MI      & Weighted F1 $\uparrow$ & 60.62 $\pm$ 0.95 & 43.10 $\pm$ 1.36 & 48.95 $\pm$ 1.87 & 49.95 $\pm$ 4.55 & \textbf{60.93 $\pm$ 0.95} \\
        SHU-MI            & AUROC $\uparrow$   & 64.31 $\pm$ 0.82 & 66.16 $\pm$ 1.67 & 64.58 $\pm$ 1.01 & 65.70 $\pm$ 1.50  & \textbf{66.48 $\pm$ 0.61} \\
        SEED-V     & Weighted F1 $\uparrow$   & 26.08 $\pm$ 1.80 & 24.38 $\pm$ 2.75 & 26.32 $\pm$ 1.04 & 22.74 $\pm$ 1.67 & \textbf{28.16 $\pm$ 0.71} \\
        STEW     & Weighted F1 $\uparrow$   & 49.15 $\pm$ 6.56 & 46.87 $\pm$ 2.63 & 49.09 $\pm$ 3.15 & 46.77 $\pm$ 1.73 & \textbf{52.90 $\pm$ 2.56} \\
        SEED-VIG & NRMSE $\downarrow$ & 1.01 $\pm$ 0.03 & 0.95 $\pm$ 0.04 & 0.94 $\pm$ 0.05 & 0.99 $\pm$ 0.04 & \textbf{0.93 $\pm$ 0.03}  \\
        TUAB   & AUROC $\uparrow$ & 88.97 $\pm$ 0.33 & 88.56 $\pm$ 0.74 & \textbf{89.36 $\pm$ 0.46} & 88.30 $\pm$ 0.40 & 89.30 $\pm$ 0.25  \\
        ISRUC & Weighted F1 $\uparrow$   & 77.19 $\pm$ 1.05 & 77.90 $\pm$ 1.46 & \textbf{78.93 $\pm$ 0.72} & 77.29 $\pm$ 0.83 & 78.23 $\pm$ 0.59 \\
        Mumtaz   & AUROC $\uparrow$   & 97.81 $\pm$ 0.83 & \textbf{99.01 $\pm$ 0.90} & 98.69 $\pm$ 0.27 & 95.74 $\pm$ 0.77 & 98.23 $\pm$ 0.17  \\
        Mental Arithmetic & AUROC $\uparrow$ & \textbf{75.66 $\pm$ 2.77} & 70.88 $\pm$ 6.77 & 72.23 $\pm$ 3.16 & 62.84 $\pm$ 2.75 & 68.97 $\pm$ 1.71  \\
        \bottomrule
    \end{tabular}%
    }
\end{table*}

\begin{table*}[!hbtp]
    \centering
    \caption{Performance on STEW (3-class mental workload classification) and SEED-VIG (regression on vigilance estimates).}
    \label{tab:stew_seedvig}
    \resizebox{\textwidth}{!}{%
    \footnotesize
    \setlength{\tabcolsep}{5pt}
    \renewcommand{\arraystretch}{1.05}
    \begin{tabular}{l c c c c c c c}
        \toprule
        \multirow{2}{*}{\textbf{Methods}}
          & \multirow{2}{*}{\textbf{Model Size}}
          & \multicolumn{3}{c}{\textbf{STEW, 3-class}}
          & \multicolumn{3}{c}{\textbf{SEED-VIG, Regression}} \\
        \cmidrule(lr){3-5}\cmidrule(lr){6-8}
         &
         & \textbf{Balanced Acc.} $\uparrow$ & \textbf{Cohen's $\kappa$} $\uparrow$ & \textbf{Weighted F1} $\uparrow$
         & \textbf{NRMSE} $\downarrow$ & \textbf{RMSE} $\downarrow$ & \textbf{Pearson} $\uparrow$ \\
        \midrule
        EEGNet                  & 0.003M & 49.58 $\pm$ 2.37 & 0.2556 $\pm$ 0.0321 & 48.99 $\pm$ 2.56 & 1.0753 $\pm$ 0.0375 & 0.1884 $\pm$ 0.0066 & 0.2741 $\pm$ 0.0430 \\
        EEGConformer            & 0.55M  & 49.97 $\pm$ 7.06 & 0.2910 $\pm$ 0.0683 & 49.15 $\pm$ 6.56 & 1.5028 $\pm$ 0.1056 & 0.2633 $\pm$ 0.0185 & 0.2837 $\pm$ 0.0595 \\
        SpaRCNet                & 0.79M  & 45.49 $\pm$ 3.99 & 0.1952 $\pm$ 0.0564 & 46.14 $\pm$ 3.88 & 1.3257 $\pm$ 0.1841 & 0.2323 $\pm$ 0.0323 & 0.2658 $\pm$ 0.0932 \\
        CNN-Transformer         & 1.6M   & 50.89 $\pm$ 2.01 & 0.2643 $\pm$ 0.0366 & 48.04 $\pm$ 1.35 & 1.0095 $\pm$ 0.0306 & 0.1772 $\pm$ 0.0056 & 0.4284 $\pm$ 0.0417 \\
        ContraWR                & 3.2M   & 50.47 $\pm$ 3.58 & 0.2630 $\pm$ 0.0312 & 48.69 $\pm$ 4.79 & 1.0740 $\pm$ 0.0645 & 0.1882 $\pm$ 0.0113 & 0.4083 $\pm$ 0.0358 \\
        ST-Transformer          & 2.4M   & 47.81 $\pm$ 4.59 & 0.2434 $\pm$ 0.0779 & 46.04 $\pm$ 4.03 & 1.1680 $\pm$ 0.1290 & 0.2047 $\pm$ 0.0226 & 0.3108 $\pm$ 0.0336 \\
        FFCL                    & 3.5M   & 47.91 $\pm$ 4.16 & 0.2397 $\pm$ 0.0843 & 45.89 $\pm$ 6.48 & 1.0587 $\pm$ 0.0771 & 0.1855 $\pm$ 0.0135 & 0.3751 $\pm$ 0.0331 \\
        \addlinespace[2pt]
        \midrule[0.6pt]
        \addlinespace[1pt]
        BIOT                    & 3.2M   & 47.61 $\pm$ 1.75 & 0.2367 $\pm$ 0.0263 & 46.87 $\pm$ 2.63 & 0.9528 $\pm$ 0.0426 & 0.1670 $\pm$ 0.0075 & 0.4253 $\pm$ 0.0683 \\
        CBraMod                 & 4.0M   & 48.30 $\pm$ 3.44 & 0.2498 $\pm$ 0.0550 & 49.09 $\pm$ 3.15 & 0.9369 $\pm$ 0.0461 & 0.1642 $\pm$ 0.0181 & \textbf{0.5189 $\pm$ 0.0307} \\
        LaBraM                  & 5.8M   & 49.42 $\pm$ 1.45 & 0.2530 $\pm$ 0.0107 & 46.77 $\pm$ 1.73 & 0.9895 $\pm$ 0.0370 & 0.1734 $\pm$ 0.0065 & 0.4508 $\pm$ 0.0112 \\
        \addlinespace[2pt]
        \midrule[0.6pt]
        \addlinespace[1pt]
        CEReBrO                 & 2.4M   & 52.15 $\pm$ 3.58 & 0.2949 $\pm$ 0.0503 & 49.80 $\pm$ 3.74 & 0.9328 $\pm$ 0.0237 & 0.1636 $\pm$ 0.0142 & 0.4501 $\pm$ 0.0366 \\
        S-CEReBrO               & 2.4M   & \textbf{55.29 $\pm$ 1.66} & \textbf{0.3162 $\pm$ 0.0267} & \textbf{52.90 $\pm$ 2.56} & \textbf{0.9284 $\pm$ 0.0316} & \textbf{0.1522 $\pm$ 0.0228} & 0.4710 $\pm$ 0.0222 \\
        \bottomrule
    \end{tabular}
    }
\end{table*}

\subsection{Computational Complexity and Scaling Analysis}
\label{subsec:attention_efficiency}
\begin{figure*}[!hbtp]
    \centering
    \begin{subfigure}[b]{0.49\textwidth}
        \centering
        \includegraphics[width=\textwidth]{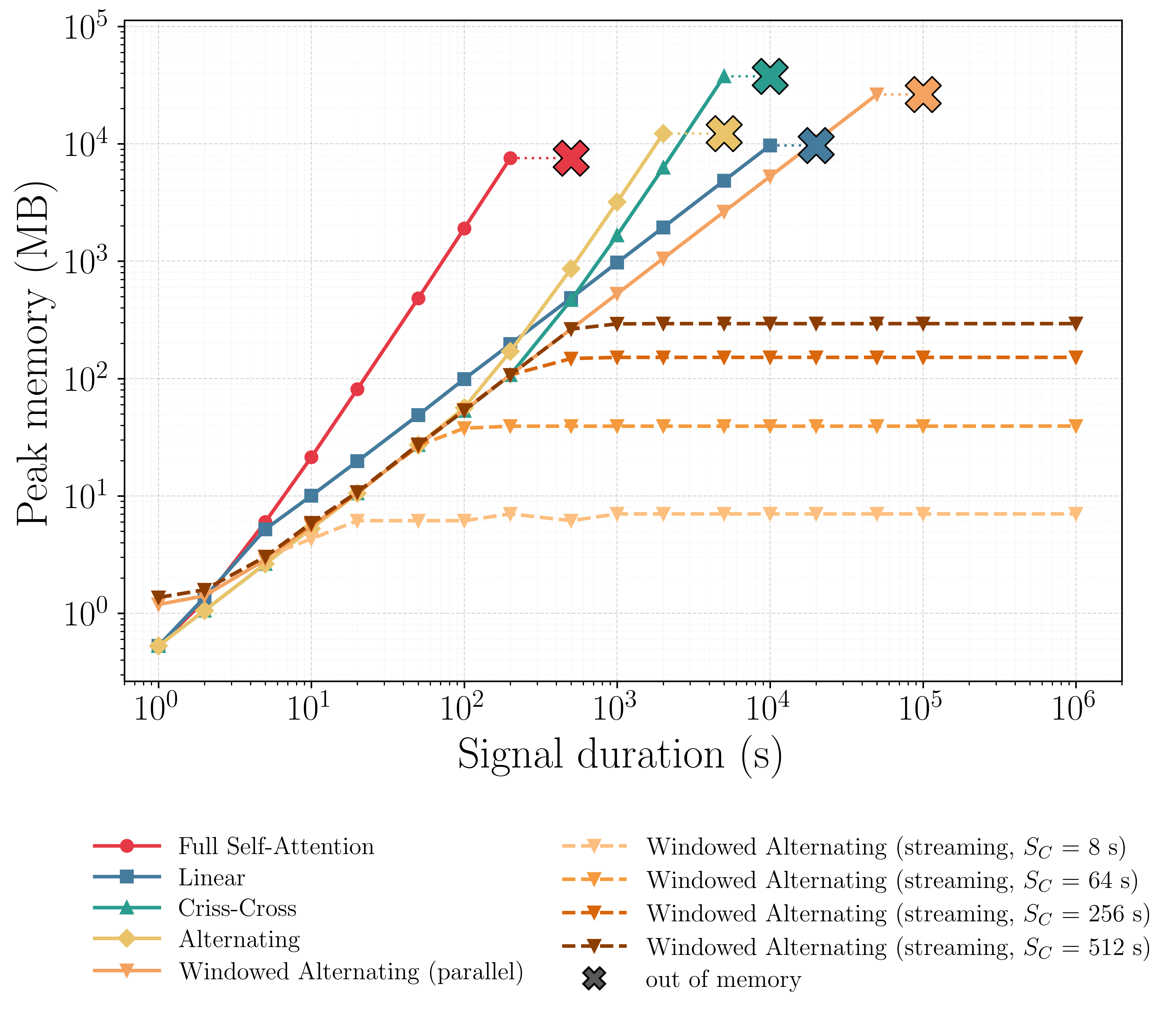}
        \caption{Memory usage vs. signal duration}
        \label{fig:attention_memory}
    \end{subfigure}
    \hfill
    \begin{subfigure}[b]{0.49\textwidth}
        \centering
        \includegraphics[width=\textwidth]{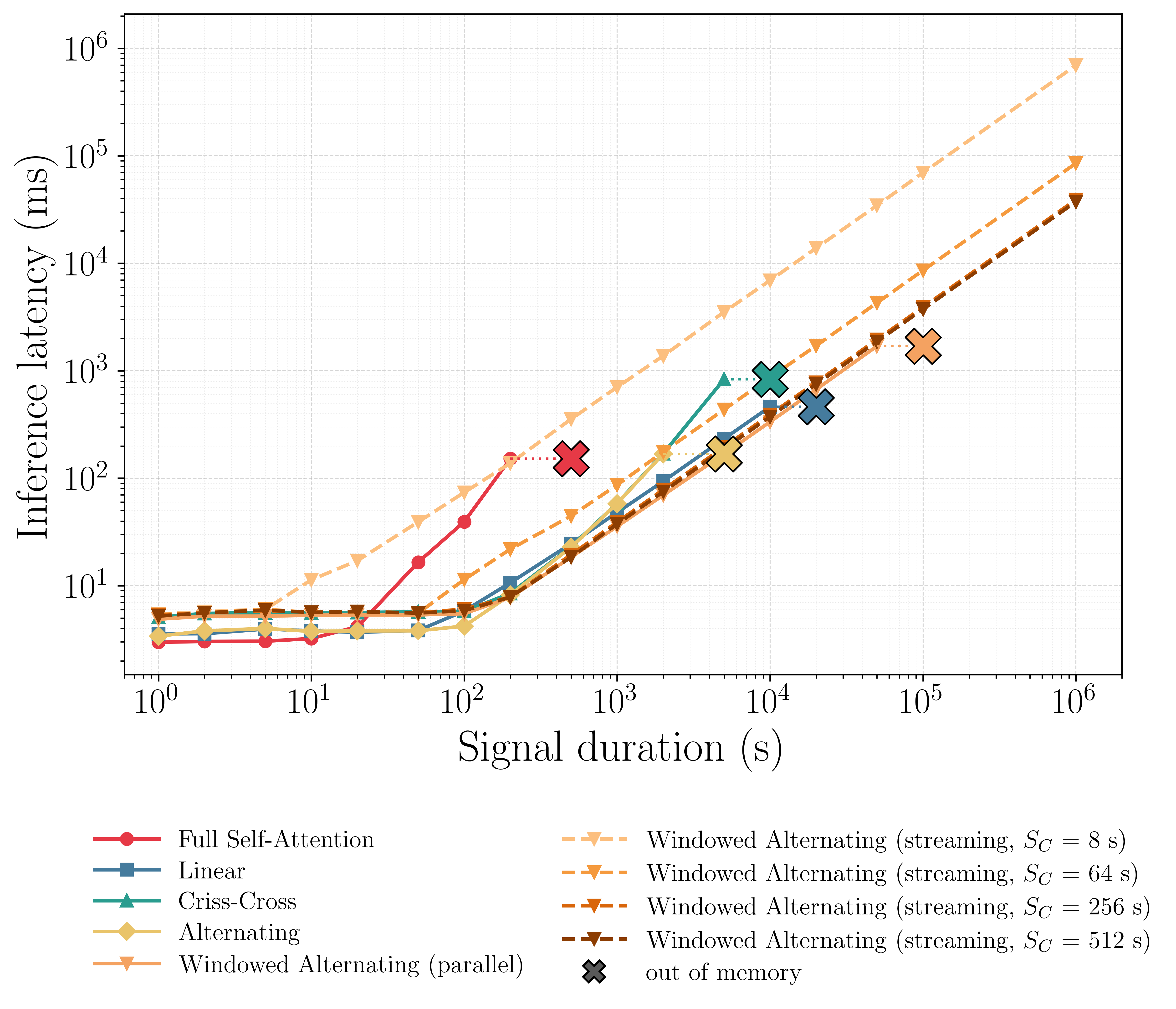}
        \caption{Inference latency vs. signal duration}
        \label{fig:attention_runtime}
    \end{subfigure}
    \caption{Peak memory and latency vs. signal duration $T$. Windowed alternating attention (parallel) processes the full signal simultaneously, scaling memory and runtime linearly. Streaming mode processes sequential chunks, bounding memory independently of $T$ ($S_C$ denotes chunks in the computational buffer). Crosses mark "Out of Memory" on a 40GB A100 GPU.}
    \label{fig:attention_comparison}
\end{figure*}

We conduct scaling experiments to evaluate windowed alternating attention against established attention baselines for long-form EEG monitoring, with results illustrated in \Cref{fig:attention_comparison}. \edit{At $T=15,000$ seconds (4 hours), windowed alternating attention requires 55\% of the memory and achieves 2.1$\times$ higher inference throughput than the second-best method (linear attention)}. Moreover, windowed alternating attention is the only mechanism to avoid a memory limit, successfully processing signals up to \edit{50,000 seconds (14 hours); a duration 100$\times$ longer than for full self-attention, 10$\times$ longer than for alternating attention, and 3$\times$ longer than for linear attention}.

\section{Conclusion}
\label{sec:conclusion}

We propose S-CEReBrO, a foundation model that mitigates memory bottlenecks in continuous EEG monitoring. \textit{Windowed Alternating Attention} maintains a constant resident memory independent of the signal length, \edit{processing signals up to 100$\times$ longer than self-attention and 3$\times$ longer than low-rank linear attention. This localized spatiotemporal inductive bias drives efficiency and accuracy, achieving state-of-the-art performance on 7 of 11 tasks with up to 60\% fewer parameters than existing foundation models. Ultimately, S-CEReBrO paves the way towards high-fidelity, continuous, real-time EEG analysis. Future work will focus on developing long-form benchmarks to evaluate hour-scale streaming, and improving sample efficiency for better adaptation to exceptionally small datasets}.

\subsubsection*{Acknowledgements.} This work was supported by a grant from the Swiss National Supercomputing Centre (CSCS) under project IDs lp12 and lp160 on Alps, the ETH Future Computing Laboratory (EFCL), and a donation from Huawei Technologies.

\subsubsection*{Disclosure of Interests.} The authors have no competing interests to declare that
are relevant to the content of this article.

%
%
%
\bibliographystyle{splncs04}
\bibliography{Paper-6624}

@article{SEED-IV,
	title        = {{EmotionMeter: A Multimodal Framework for Recognizing Human Emotions}},
	author       = {Wei-Long Zheng and Wei Liu and Yifei Lu and Bao-Liang Lu and Andrzej Cichocki},
	year         = 2018,
	journal      = {IEEE Transactions on Cybernetics},
	pages        = {1--13},
	issn         = {2168-2267}
}

@article{SEED-FRA,
	title        = {{Identifying Similarities and Differences in Emotion Recognition with EEG and Eye Movements among Chinese, German, and French People}},
	author       = {Wei Liu and others},
	year         = 2022,
	journal      = {Journal of Neural Engineering},
	volume       = 19,
	number       = 2,
	pages        = {026012},
}

@article{EEGConformer,
	title        = {{EEGConformer: Convolutional Transformer for EEG Decoding and Visualization}},
	author       = {Song, Yonghao and Zheng, Qingqing and Liu, Bingchuan and Gao, Xiaorong},
	year         = 2023,
	journal      = {IEEE Transactions on Neural Systems and Rehabilitation Engineering},
	volume       = 31,
	number       = {},
	pages        = {710--719}
}

@misc{BIOT,
	title        = {{BIOT: Cross-data Biosignal Learning in the Wild}},
	author       = {Chaoqi Yang and M. Brandon Westover and Jimeng Sun},
	year         = 2023,
	eprint       = {2305.10351},
	archiveprefix = {arXiv},
	primaryclass = {eess.SP}
}

@misc{ContraWR,
	title        = {{Self-supervised EEG Representation Learning for Automatic Sleep Staging}},
	author       = {Chaoqi Yang and others},
	year         = 2023,
	eprint       = {2110.15278},
	archiveprefix = {arXiv},
	primaryclass = {eess.SP}
}

@inproceedings{LaBraM,
	title        = {{Large Brain Model for Learning Generic Representations with Tremendous EEG Data in BCI}},
	author       = {Weibang Jiang and Liming Zhao and Bao-liang Lu},
	year         = 2024,
	booktitle    = {The Twelfth International Conference on Learning Representations},
}

@inproceedings{patchtst,
	title        = {{A Time Series is Worth 64 Words:  Long-term Forecasting with Transformers}},
	author       = {Yuqi Nie and others},
	year         = 2023,
	booktitle    = {The Eleventh International Conference on Learning Representations},
}

@inproceedings{CBraMod,
	title        = {{{CB}raMod: A Criss-Cross Brain Foundation Model for {EEG} Decoding}},
	author       = {Jiquan Wang and Sha Zhao and Zhiling Luo and Yangxuan Zhou and Haiteng Jiang and Shijian Li and Tao Li and Gang Pan},
	year         = 2025,
	booktitle    = {The Thirteenth International Conference on Learning Representations},
}

@article{Neonate,
	title        = {{A dataset of neonatal EEG recordings with seizure annotations}},
	author       = {Stevenson, Nathan J and Tapani, Karoliina and Lauronen, Leena and Vanhatalo, Sampsa},
	year         = 2019,
	month        = {Mar},
	day          = 5,
	journal      = {Scientific Data},
	publisher    = {Nature Publishing Group},
	volume       = 6,
	pages        = 190039,
	pmid         = 30835259,
	pmcid        = {PMC6400100}
}

@article{obeid2016temple,
	title        = {{The Temple University Hospital EEG Data Corpus}},
	author       = {Obeid, Iyad and Picone, Joseph},
	year         = 2016,
	journal      = {Frontiers in Neuroscience},
	publisher    = {Frontiers},
	volume       = 10,
	pages        = 196,
}

@article{EEGNet,
	title        = {{EEGNet: a compact convolutional neural network for EEG-based brain–computer interfaces}},
	author       = {Lawhern, Vernon J and Solon, Amelia J and Waytowich, Nicholas R and Gordon, Stephen M and Hung, Chou P and Lance, Brent J},
	year         = 2018,
	month        = jul,
	journal      = {Journal of Neural Engineering},
	publisher    = {IOP Publishing},
	volume       = 15,
	number       = 5,
	pages        = {056013},
	issn         = {1741-2552},
}

@article{SEED,
	title        = {{Investigating critical frequency bands and channels for EEG-based emotion recognition with deep neural networks}},
	author       = {Zheng, Wei-Long and Lu, Bao-Liang},
	year         = 2015,
	journal      = {IEEE Transactions on Autonomous Mental Development},
	publisher    = {IEEE},
	volume       = 7,
	number       = 3,
	pages        = {162--175},
}

@article{SleepEDFx,
	title        = {{Analysis of a sleep-dependent neuronal feedback loop: the slow-wave microcontinuity of the EEG}},
	author       = {Kemp, B and others},
	year         = 2000,
	journal      = {IEEE Transactions on Biomedical Engineering},
	publisher    = {IEEE},
	volume       = 47,
	number       = 9,
	pages        = {1185--1194},
}

@article{BCI-NER,
  author    = {Perrin, M. and Maby, E. and Daligault, S. and Bertrand, O. and Mattout, J.},
  title     = {{Objective and subjective evaluation of online error correction during P300-based spelling}},
  journal   = {Advances in Human-Computer Interaction},
  year      = {2012},
  volume    = {2012},
  number    = {4}
}

@misc{BOAS,
  author = {Eduardo López-Larraz and others},
  title = {{Bitbrain Open Access Sleep Dataset}},
  year = {2025},
  publisher = {OpenNeuro}
}

@article{SPaRCNet,
	title        = {{Development of Expert-Level Classification of Seizures and Rhythmic and Periodic Patterns During EEG Interpretation}},
	author       = {Jing, J. and others},
	year         = 2023,
	month        = apr,
	journal      = {Neurology},
	volume       = 100,
	number       = 17,
	pages        = {e1750-e1762},
	note         = {PMID: 36878708; PMCID: PMC10136013}
}

@misc{CNN-Transformer,
	title        = {{Transformer Convolutional Neural Networks for Automated Artifact Detection in Scalp EEG}},
	author       = {Wei Yan Peh and others},
	year         = 2022,
	eprint       = {2208.02405},
	archiveprefix = {arXiv},
	primaryclass = {eess.SP}
}

@article{FFCL,
	title        = {{Motor imagery EEG classification algorithm based on CNN-LSTM feature fusion network}},
	author       = {Hongli Li and Man Ding and Ronghua Zhang and Chunbo Xiu},
	year         = 2022,
	month        = {February},
	journal      = {Biomedical Signal Processing and Control},
	volume       = 72,
	pages        = 103342,
	issn         = {1746-8094},
	part         = {A}
}

@misc{ST-Transformer,
	title        = {{Transformer-based Spatial-Temporal Feature Learning for EEG Decoding}},
	author       = {Yonghao Song and Xueyu Jia and Lie Yang and Longhan Xie},
	year         = 2021,
	eprint       = {2106.11170},
	archiveprefix = {arXiv},
	primaryclass = {eess.SP}
}

@misc{wang2020linformer,
	title        = {{Linformer: Self-attention with linear complexity}},
    eprint={2006.04768},
    archivePrefix={arXiv},
	author       = {Wang, Sinong and others},
	year         = 2020,
}

@article{SEED-V,
	title        = {{Comparing recognition performance and robustness of multimodal deep learning models for multimodal emotion recognition}},
	author       = {Liu, Wei and Qiu, Jie-Lin and Zheng, Wei-Long and Lu, Bao-Liang},
	year         = 2021,
	journal      = {IEEE Transactions on Cognitive and Developmental Systems},
	volume       = 14,
	number       = 2,
	pages        = {715--729}
}

@article{PhysioNet-MI,
	title        = {{BCI2000: A general-purpose brain-computer interface (BCI) system}},
	author       = {Schalk, Gerwin and McFarland, Dennis J. and Hinterberger, Thilo and Birbaumer, Niels and Wolpaw, Jonathan R.},
	year         = 2004,
	journal      = {IEEE Transactions on Biomedical Engineering},
	volume       = 51,
	number       = 6,
	pages        = {1034--1043}
}

@article{SHU-MI,
	title        = {{A large EEG dataset for studying cross-session variability in motor imagery brain-computer interface}},
	author       = {Ma, Jun and Yang, Banghua and Qiu, Wenzheng and Li, Yunzhe and Gao, Shouwei and Xia, Xinxing},
	year         = 2022,
	journal      = {Scientific Data},
	volume       = 9,
	number       = 1,
	pages        = 531
}

@article{ISRUC,
	title        = {{ISRUC-sleep: A comprehensive public dataset for sleep researchers}},
	author       = {Khalighi, Sirvan and Sousa, Teresa and Santos, Jos{\'e} Moutinho and Nunes, Urbano},
	year         = 2016,
	journal      = {Computer Methods and Programs in Biomedicine},
	volume       = 124,
	pages        = {180--192}
}

@phdthesis{CHB-MIT,
	title        = {Application of Machine Learning to Epileptic Seizure Onset Detection and Treatment},
	author       = {Shoeb, Ali Hossam},
	year         = 2009,
	school       = {Massachusetts Institute of Technology}
}

@misc{Mumtaz2016,
	title        = {{MDD Patients and Healthy Controls {EEG} Data}},
	author       = {Mumtaz, Wajid},
	year         = 2016,
	note         = {Dataset. doi:10.6084/m9.figshare.4244171.v2},
    }

@article{SEED-VIG,
  author = {Zheng, W. L. and Lu, B. L.},
  title = {{A multimodal approach to estimating vigilance using EEG and forehead EOG}},
  journal = {Journal of Neural Engineering},
  volume = {14},
  number = {2},
  pages = {026017},
  year = {2017},
  month = {apr},
}

@article{MentalArithmetic,
	title        = {Electroencephalograms during mental arithmetic task performance},
	author       = {Zyma, Igor and others},
	year         = 2019,
	journal      = {Data},
	volume       = 4,
	number       = 1,
	pages        = 14
}

@article{Kobler2018,
  author    = {Reinmar J. Kobler and Andreea I. Sburlea and Gernot R. M{\"u}ller-Putz},
  title     = {{Tuning Characteristics of Low-Frequency EEG to Positions and Velocities in Visuomotor and Oculomotor Tracking Tasks}},
  journal   = {Scientific Reports},
  year      = {2018},
}

@misc{STEW,
  author       = {Lim, Wei Lun and Sourina, Olga and Wang, Lipo},
  title        = {{STEW: Simultaneous Task EEG Workload Dataset}},
  year         = {2018},
  publisher    = {IEEE Dataport},
}

@inproceedings{
LUNA,
title={{LUNA: Efficient and Topology-Agnostic Foundation Model for {EEG} Signal Analysis}},
author={Berkay D{\"o}ner and Thorir Mar Ingolfsson and Luca Benini and Yawei Li},
booktitle={The Thirty-ninth Annual Conference on Neural Information Processing Systems},
year={2025},
}

@misc{CEReBrO,
      title={{CEReBrO: Compact Encoder for Representations of Brain Oscillations Using Efficient Alternating Attention}}, 
      author={Alexandru Dimofte and Glenn Anta Bucagu and Thorir Mar Ingolfsson and Xiaying Wang and Andrea Cossettini and Luca Benini and Yawei Li},
      year={2025},
      eprint={2501.10885},
      archivePrefix={arXiv},
      primaryClass={cs.LG},
      howpublished={arXiv:2501.10885 [cs.LG]},
}

@misc{longformer,
      title={Longformer: The Long-Document Transformer}, 
      author={Iz Beltagy and Matthew E. Peters and Arman Cohan},
      year={2020},
      eprint={2004.05150},
      archivePrefix={arXiv},
      primaryClass={cs.CL}, 
}

\end{document}